# Leveraging Activity Recognition to Enable Protective Behavior Detection in Continuous Data

Chongyang Wang

University College London, United Kingdom, chongyang.wang.17@ucl.ac.uk

Yuan Gao

Uppsala University, Sweden, alex.yuan.gao@it.uu.se

Akhil Mathur

Nokia Bell Labs, United Kingdom, akhil.mathur@nokia-bell-labs.com

Amanda C. De C. Williams

University College London, United Kingdom, amanda.williams@ucl.ac.uk

Nicholas D. Lane

University of Cambridge, United Kingdom, ndl32@cam.ac.uk

Nadia Bianchi-Berthouze

University College London, United Kingdom, nadia.berthouze@ucl.ac.uk

Protective behavior exhibited by people with chronic pain (CP) during physical activities is the key to understanding their physical and emotional states. Existing automatic protective behavior detection (PBD) methods rely on pre-segmentation of activities predefined by users. However, in real life, people perform activities casually. Therefore, where those activities present difficulties for people with chronic pain, technology-enabled support should be delivered continuously and automatically adapted to activity type and occurrence of protective behavior. Hence, to facilitate ubiquitous CP management, it becomes critical to enable accurate PBD over continuous data. In this paper, we propose to integrate human activity recognition (HAR) with PBD via a novel hierarchical HAR-PBD architecture comprising graph-convolution and long short-term memory (GC-LSTM) networks, and alleviate class imbalances using a class-balanced focal categorical-cross-entropy (CFCC) loss. Through in-depth evaluation of the approach using a CP patients' dataset, we show that the leveraging of HAR, GC-LSTM networks, and CFCC loss leads to clear increase in PBD performance against the baseline (macro F1 score of 0.81 vs. 0.66 and precision-recall area-under-the-curve (PR-AUC) of 0.60 vs. 0.44). We conclude by discussing possible use cases of the hierarchical architecture in CP management and beyond. We also discuss current limitations and ways forward.

## 1 INTRODUCTION

Chronic pain is a prevalent condition in ~30.7% of adults in the US [1]. People with chronic musculoskeletal pain (a common type of chronic pain) exhibit **protective behavior** (guarding, hesitation, the use of support, abrupt motion, and rubbing) during physical activity [33], providing important information about their physical and emotional states and ability to manage their condition [34, 35]. In clinical settings, physiotherapists observe protective behavior and respond with feedback, movement suggestions, and therapeutic interventions [5]. This tailored support is critical to incrementally build patients' self-efficacy and maintain their engagement in physical activity [36]. However, such support is expensive and only available to few people with CP. In addition, behavior in the clinic [68] is a narrow sample of the physical and psychological capabilities required for everyday physical functioning. Maintaining self-management is hard and people often disengage, thereby losing valued activities including social involvement [34]. To prevent disengagement, observation and personalized support need to extend beyond the clinical context [37].

Ubiquitous sensing and computing technology offer new opportunities to provide such support to people with CP. Patients describe technology capable of protective behavior detection (PBD) as a 'second pair of eyes', increasing their awareness and helping application of pain management strategies learned in the clinic [62]. In [79], patients and physiotherapists discussed how such technology could help patients to better control activity pacing and breathing when protective behavior is detected. The technology may also, e.g. replicate physiotherapists' advice on chair height if the patient has difficulties sitting down or standing up. These studies also show that awareness of habitual protective behavior can help reduce it (e.g., reminding the person to bend the trunk as they stand up from a chair). In addition to providing personalized feedback, such technology can be adopted to evaluate the effect of clinical interventions [65].

The first step in building a ubiquitous technology to help people with CP in their everyday lives is to enable **continuous** PBD during diverse functional activities. To date, the focus has been on PBD in specific exercises where the activity being performed is known in advance. Interesting PBD results are only achieved within **pre-segmented** activity instances [25, 40, 64]. However, pre-segmentation is not feasible for everyday (functional) activities. In this paper, we aim to address these problems by approaching continuous PBD with continuous recognition of the activity (HAR) in process. We propose a novel hierarchical HAR-PBD architecture, where the activity type when recognized is continuously leveraged to build **activity-informed** input for concurrent PBD.

To investigate the efficacy of our approach, we use the fully-annotated EmoPain dataset [32]. The dataset comprises full-body movement data captured from CP and healthy participants during sequences of movements reflecting everyday activities. We refer to these as activities-of-interest (AoIs) since they were chosen by physiotherapists as particularly demanding for people with CP and likely to trigger protective behavior. While this dataset was not collected in the wild, participants performed each activity without instruction, and transitions between AoIs created further noise typical of in-the-wild data collection. During transition periods, participants could rest according to their needs or enjoy casual movements such as stretching, walking, and self-preparations. An illustration of a complete activity sequence of one CP participant with the protective behavior annotation is shown in Figure 1. Evaluation shows that the activity information noticeably improves the PBD performance in such continuous data, achieving macro F1 score of 0.73 and PR-AUC of 0.52 in comparison with the baseline method without such information (macro F1 score of 0.66 and PR-AUC of 0.44). By alleviating class imbalances with a class-balanced focal categorical-cross-entropy (CFCC) loss [59, 60], PBD performance is further improved, achieving macro F1 score of 0.81 and PR-AUC of 0.60. Our contributions are four-fold:

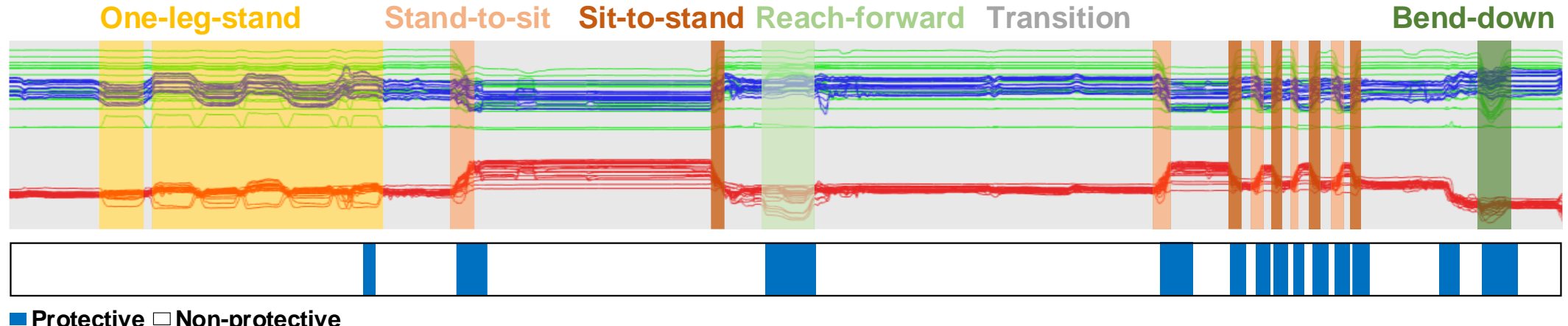

Figure 1: An example of the full data sequence from a CP participant, comprising AoIs and transitions. Lines are red, green, and blue for the x, y, and z coordinates data respectively. Protective behavior labels (majority-voted) are shown below the sequence.

1. For the first time, continuous detection of protective behavior is studied using full data sequences of CP patients. Previously, continuous PBD was only established on pre-segmented activity instances;
2. A novel hierarchical HAR-PBD architecture is designed to leverage activity recognition to enable detection of protective behavior (i.e., movement behavior driven by emotional variables) in continuous data sequences. Protective behavior was investigated in the past without leveraging its activity background;
3. Graph convolution (GC) [63] and long short-term memory (LSTM) [61] layers are combined to model the body-worn inertial measurement units (IMUs) data for PBD, while in the past only convolutional neural networks (CNNs) [66] and LSTMs were applied. Although the concept of combining GC and LSTM exists in computer vision-based studies, it is adopted for the first time to show advantage of graph representation in the context of emotional behavior across activities. A loss function referred to as CFCC loss is also employed to alleviate class imbalances of continuous data;
4. Comprehensive experiments and analyses using data collected from both CP and healthy participants. Various training strategies of the proposed hierarchical architecture are explored, and an analysis of simulating fewer IMUs demonstrates the applicability and efficacy of our method on smaller sensor sets.

## 2 BACKGROUND, MOTIVATIONS AND RELATED WORKS

Our proposed hierarchical HAR-PBD architecture comprises two main modules: one for activity recognition and another for protective behavior detection. Here we summarize the literature related to pain-, fear-, and anxiety-induced movement behavior detection and human activity recognition, while reveal the motivations of this work.

### 2.1 Affective Movement Behavior Detection

Pain, fear, and anxiety are expressed not only by the face, but also by altered body movements [2, 80]. The automatic detection of affective bodily expressions is a growing area of research in the affective computing community [3, 74]. While bodily expressions of emotion were previously studied in isolation, the focus is now on real-life data. Due to the technical challenges, most studies still use static situations (e.g. during a consultation interview with a therapist [6]) or the type of activity is constant throughout (e.g. the detection of pain and anxiety in game-based physical rehabilitation [71, 78]). Bodily expression is also used to inform healthcare applications, e.g. for detection of depression [6], oral hygiene [7], and perinatal assessment for stroke [8]. Typically, these scenarios only require the tracking of few body parts without fine-grained analysis of full-body movement.

Automatic detection of continuous affective behavior across different daily activities is still rare. For example, [4] explored the detection of bodily expressions of reflective thinking in the context of diverse full-body mathematical games. While this study developed activity-independent models over continuous data sequences, their proposed LSTM-based architecture needs to be trained on pre-segmented affective events (e.g. when the child expresses the states of interest vs. other states). Recently, studies very relevant to ours have attempted to detect protective behavior across different activities. On the EmoPain dataset [32], researchers have shown that the use of LSTM-based architectures facilitates activity-independent PBD with improved performances. Interesting results are seen in [40] and [64], where the stacked-LSTM network and body attention network (BANet) were proposed to conduct traversal and local processing of body movement data respectively. Although the model is activity-independent and functions across different activity types, continuous detection was constrained only within pre-segmented AoIs. The relationship between the type of activity and protective behavior is not leveraged in the modeling. The attention mechanism used in BANet only focuses on identifying the most relevant body segments but does not directly leverage such relationship.

As such, how to enable continuous PBD along a sequence of activities remains an open challenge. The high variability of protective behavior across people exhibited within the same activity type [64] also calls for a better approach to extract useful information from the full-body movement data.

### 2.2 Human Activity Recognition

The modeling of body movement has gone through extensive development in the context of human activity recognition (HAR). The majority of HAR research focuses on classifying the type of activity a person is engaged in by using data from wearable sensors [9-16] or skeleton data from visual motion-capture (MoCap) systems [17, 18]. The preference for wearable sensors vs. visual systems rests on the limits to mobility imposed by the application.

HAR with vision- and sensor-based data has evolved quickly in the past few years, especially for the perspective of data processing strategies. Initially, data was processed in a traversal manner, where acceleration, orientation, or joint coordinates were treated as temporal multi-dimensional sequences. As a result, efforts were dedicated to feature engineering [38, 39] and basic neural networks [9-11] e.g. LSTM networks [61], to address the temporal aspects of body movement data. Later, various studies started to exploit the spatial configuration of the sensor/joint network. For instance, several data representations considered the relative relationships between sensors/joints [41-43], with network architectures designed to enable local processing of movement dynamics [12, 13, 44-46, 64]. Performance improvements achieved by these methods suggest that body configuration information is important for activity recognition. More recently, the re-introduction of graph convolution network (GCN) [54, 55] offers a new method for HAR. One reason for the successful use of GCN on skeleton-based movement data [17, 18, 24, 47-49] is that the human body can be naturally presented as a non-directed graph. Graph representation helps a model learn the biomechanical relationships between body segments without imposing knowledge about specific activities of interest. Noticeable improvements are seen on several benchmark datasets (e.g. NUS RGB+ [19] and Kinetics [20]), achieved by using GCNs.

Whilst the concept of body configuration is very much leveraged in vision-based HAR systems, enabled by the full-body MoCap therein, it is not the case for ubiquitous sensor-based HAR or movement behavior detection. The sensor-based HAR literature has focused on using a small set of sensors to classify activity, with each study examining specific activities [15] or benchmark datasets [21-23]. Using a small network of sensors also increases applicability and reduces cost in real-life deployment. However, as in the case of CP rehabilitation, critical information may not be in the movement of the main body segments involved, but in other body parts recruited to protect the body [33-36]. For example, Olugbade et al. [37] show the importance of head stiffness to indicate protective behavior during sit-to-stand-to-sit and reach-forward, although head movement is not needed to perform such activities. Psychology studies in chronic pain point to the importance of assessing movement quality rather than activity quantity. As a result, using full-body movement data (as in the EmoPain dataset) rather than a small set of sensors, to detect protective behavior across activities, is based on three arguments: i) full-body movement data is needed to capture detailed movement behavior of multiple body parts for PBD across activities; ii) patients and clinicians see benefits and opportunities that such sensing technology offers, and are open to using it [65]; iii) full-body sensing is becoming more convenient as wearable sensors are becoming smaller and integrated into clothes [76]. Moreover, we evaluate the efficacy of our method on small sensor sets at the end of this paper.

The advantage of using GCN in skeleton-based HAR, the need to model a large set of sensors, and high variability in body configuration information in PBD all suggest the importance of exploring the use of GCN in the context of protective behavior. It also brings together researches on HAR and PBD (or in general emotional movement behavior detection) that have surprisingly evolved separately, despite clearly representing activity and emotional bodily expressions that co-occur in real life with each altering the other. To the best of our knowledge, only one paper has investigated the use of GCN in bodily affective expressions [49], but considers just one task (gait) and acted emotional expressions, a much simpler (stereotypical) problem to address. As such, they explored GCN alone and do not need to address the variety of activity and class imbalances of continuous data. In this paper, we aim to use the proposed hierarchical architecture to answer the questions: is HAR beneficial to PBD in continuous data and how can these two modules be connected? For each module of our proposed architecture, graph convolution is employed to model the movement data captured by multiple IMUs per timestep. Given the success of LSTM in capturing temporal patterns of protective behavior [40, 64], LSTM layers are used to model the temporal dynamics.

## 3 THE HIERARCHICAL HAR-PBD ARCHITECTURE AND CFCC LOSS

A novel hierarchical architecture combining HAR and PBD modeling is proposed to enable PBD over continuous data sequences of activities. An overview of this architecture is presented in Figure 2. Both HAR and PBD modules receive consecutive frames as the input. These are extracted with a sliding-window from the data sequence collected with 18 IMUs. For HAR module the activity type label is used for training, whereas for PBD the protective behavior label (absence and presence) is used. In addition, the first module (HAR) aims to recognize the type of activity being performed and pass such information to the second module (PBD) that recognizes the presence or absence of protective behavior. For our main experiments, the HAR module is pre-trained with activity labels on the same folds of data during each round of leave-one-subject-out validation (LOSO) used for PBD. The weight achieving the highest activity recognition accuracy is saved. The HAR module is frozen with such pre-trained weight loaded when used in the hierarchical architecture. Therein, the activity classification output is concatenated with the same original input frame and passed to train and test the PBD module using labels of protective behavior. We use this frozen (optimal) HAR module to better understand the benefit of using the proposed hierarchical HAR-PBD architecture. Further analyses using non-frozen HAR module are reported at the end of the paper. To the best of our knowledge, this is the first implementation to leverage HAR to enable another concurrent task on the same data.

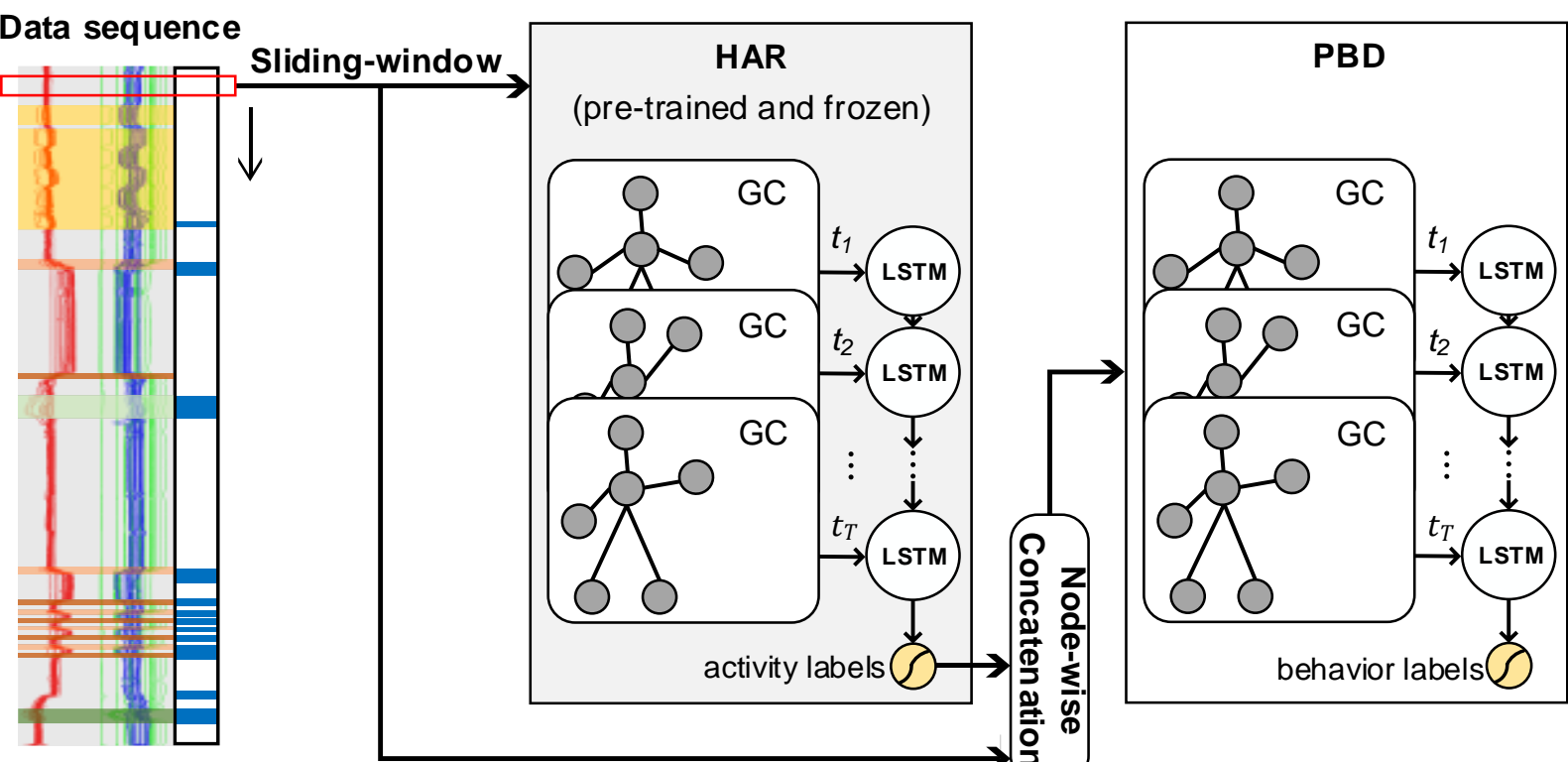

Figure 2: The proposed hierarchical HAR-PBD architecture, comprising the human activity recognition (HAR) module and protective behavior detection (PBD) module. By default, using the same data input, the HAR module is pre-trained with activity labels and frozen during training of the PBD module with behavior labels.

Both modules in our proposed architecture use a similar network comprising graph convolution and LSTM layers. The graph convolution method is used to model the body configuration information collected from 18 IMUs. Following its success in recent vision-based HAR literature, we aim to explore the contribution of graph convolution in PBD given the large variety in protective behavior exhibited by people with CP when performing each AoI. Meanwhile, LSTM is used to learn the temporal dynamics across graphs corresponding to the body movement at different timesteps, critical for both HAR and PBD (e.g. hesitation slows down movements, and fear of pain or perceived pain lead to difference in timing of body-part engagement for the same activity).

### 3.1 The GC-LSTM Network for HAR and PBD Modules

There is a variety of implementations of using graph convolution for skeleton-based movement data. Some have altered the graph convolution itself to facilitate a spatial-temporal operation [17, 47-49]. Others connect the GCN and LSTM via extra layers [18] or integrate graph convolution within the gates of each LSTM unit [24] to enable a recurrent computation across time. The performance of these approaches fluctuates on vision-based HAR benchmarks [19, 20], and they have been never applied in the context of emotional bodily behavior across different activities. For both HAR and PBD module in our proposed architecture, a network integrating GC and LSTM is used, referred to as HAR/PBD GC-LSTM. There are three considerations for the design of HAR/PBD GC-LSTM:

1. The limited size of the EmoPain dataset in comparison with popular vision-based HAR benchmarks [19, 20] that have been used to evaluate GCNs, making it difficult to adopt more complex existing implementations;
2. The need to verify if the graph representation is indeed capable of improving PBD, which requires using GCN as a way to learn data representations and removing unnecessary designs, e.g. embedding GCN into LSTM;
3. The aim to connect the HAR module with the PBD module, which requires the GC-LSTM network to tolerate the fusion of activity information and movement data at input level.

In this paper, we focus on a conceptually simple implementation that builds parallel connection between GC and LSTM layers as a basic component in our proposed architecture. Such implementation is helpful to verify the advantage of using a graph representation to model data from multiple IMUs in the context of HAR and PBD, and further facilitates a hierarchical connection between the two modules.

#### *3.1.1 Graph Input.*

A wearable motion capture suit named Animazoo IGS-190 [81] comprising 18 IMUs was used for the data collection. As provided in the EmoPain dataset [32], at each timestep, 3D coordinates of 22 body joints were calculated from the raw data stored in a Biovision Hierarchy (BVH) format. Within the BVH file, the metadata includes the skeleton proportion of the participant (e.g. the length of limbs) and position on the body each sensor was attached to. Using a Matlab MoCap toolbox [70], the approximate position of 22 body joints in the 3D space was estimated based on the metadata, the gyroscope, and accelerometer data. It is important to note that such transformation brings no prior knowledge of specific activities. It only reflects the definite position of each body joint in the 3D space. An illustration of such transformation from IMUs to positional triplets of body joints is shown in Figure 3(b).

#### *3.1.2 Graph Notation.*

A body-like graph is built to arrange each of the 22 joints to be a node connected naturally in the graph to the other joints, as shown in Figure 3(c). We denote the graph as $\mathcal{G} = (V, E)$, with a node set $V\{t, i\} = \{v_{ti} | t = 1, \dots, T; i = 1, \dots, N\}$ representing the $N$ nodes of a graph at timestep $t$ within a graph sequence of length $T$, and an edge set $E$ representing the edges connecting the nodes in this graph. Since in this work independent LSTM layers are used to learn the temporal dynamics across graphs at different timesteps, the inter-skeleton edge (usually represents the temporal dynamics) connecting consecutive graphs is not leveraged. Therefore, only the intra-skeleton edge (representing the connection of body joints) is considered with $E\{i, j\} = \{(v_{ti}, v_{tj}) | (i, j) \in B\}$, where $B$ is the set of naturally connected nodes (joints) of the human body graph. An adjacency matrix $\boldsymbol{A} \in \{0,1\}^{N \times N}$ is used to identify the edge $E$ between nodes, where $A_{i,j} = 1$ for the connected $i$-th and $j$-th nodes and 0 for disconnected ones. $\boldsymbol{A}$ stays the same for all the tasks in this work. In other words, the basic configuration of a graph is independent of time and participants, while the relative relationship between different body parts in different activities is learned during training. The identity matrix is $\boldsymbol{I_N} \in \{1\}^{N \times N}$, a diagonal matrix that represents the self-connection of each node in the graph. With the adjacency matrix $\boldsymbol{A}$ and identity matrix $\boldsymbol{I_N}$, the body configuration is represented by matrices and can be processed by neural networks. The feature of each node in a graph at timestep $t$ is stored in a feature matrix $\boldsymbol{X}_t^{\mathcal{G}} \in \mathbb{R}^{N \times \boldsymbol{3}}$. The raw feature of each node is the coordinates of the respective body joint, denoted as $\boldsymbol{X}_t^{\mathcal{G}}(v_{ti}) = [x_{ti}, y_{ti}, z_{ti}]$. The neighbor set of a node $v_{ti}$ is denoted as $\mathcal{N}(v_{ti}) = \{v_{tj} | d(v_{ti}, v_{tj}) \leq D\}$, with distance function $d(v_{ti}, v_{tj})$ accounting for the number of edges in the shortest path traveling from $v_{ti}$ to $v_{tj}$ and threshold $D$ defining the size of the neighbor set. Following previous studies using GCNs for action analysis [17, 18, 24, 47-49], we set $D = 1$ to adopt the 1-neighbor set of each node.

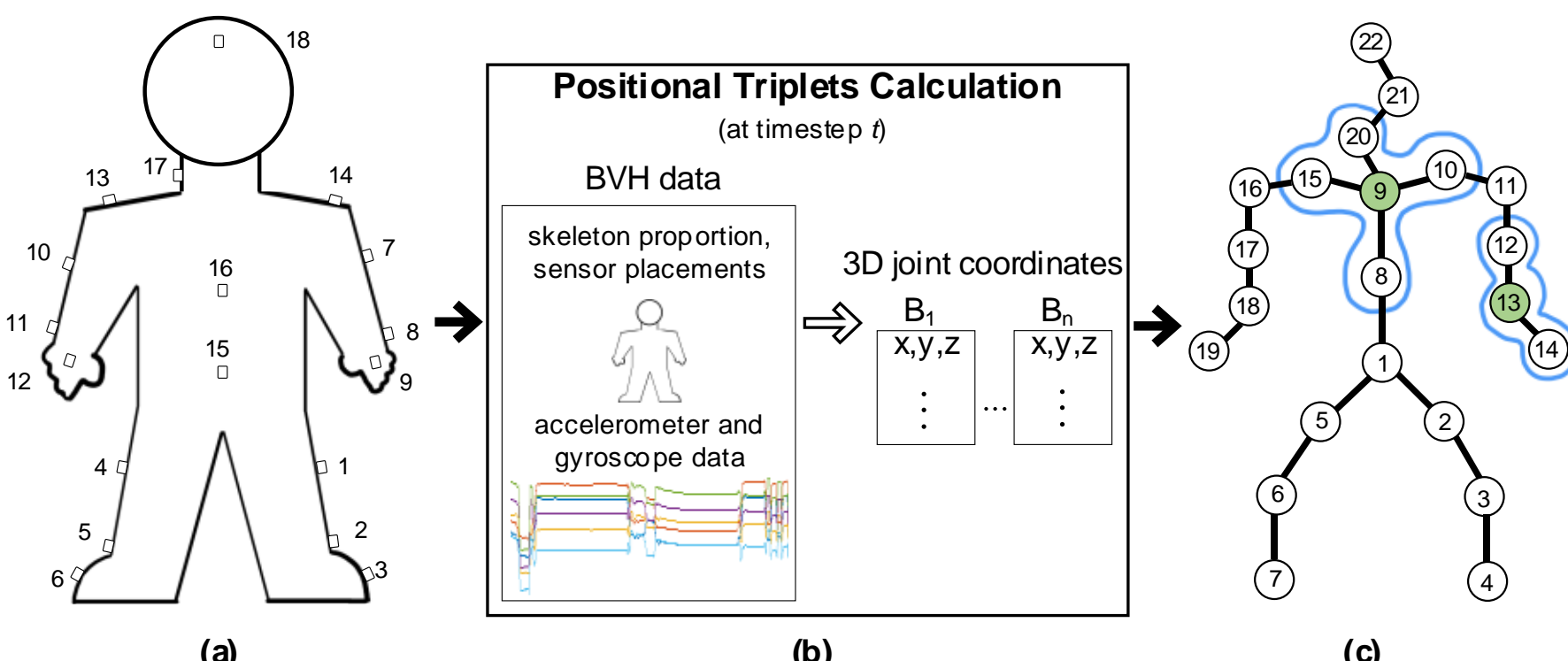

Figure 3: Illustrations of a) the placement of 18 IMUs, b) the calculation of 22 sets of 3D joint coordinates, and c) the built graph input at a single timestep, where each node represents a human body joint. The blue contour marks the neighbor set (receptive field) of the centered node in green.

#### *3.1.3 Graph Convolution.*

Basically, a graph convolution comprises two parts, one defines the way to sample data from the input graph and the other concerns assigning learnable weight to the sampled data. It should be noted that a higher-level knowledge about the subset of body parts relevant to specific activities is not manually provided in the network. Therefore, only low-level rules like sampling and weighting are defined in the graph convolution, which allows the network to develop its own understanding about the body movement. In our case, the graph convolution needs to conduct sampling on the full-body graph comprising 22 nodes.

Using the adjacency matrix $\boldsymbol{A}$ and identity matrix $\boldsymbol{I_N}$, we follow the forward-passing formula presented in [63] to implement the graph convolution used in this work as

$$\boldsymbol{f}_{out}^{GC} = \widehat{\boldsymbol{\Lambda}}^{-\frac{1}{2}}\widehat{\boldsymbol{A}}\widehat{\boldsymbol{\Lambda}}^{-\frac{1}{2}}\boldsymbol{f}_{in}^{GC}\boldsymbol{W}, \qquad (1)$$

where $\widehat{\boldsymbol{A}} = \boldsymbol{A} + \boldsymbol{I_N}$ represents the inter- and self-connection of each node, and $\widehat{\boldsymbol{\Lambda}}_{ii} = \sum_j \widehat{\boldsymbol{A}}_{ij}$ is a diagonal degree matrix of $\widehat{\boldsymbol{A}}$. Since $\widehat{\boldsymbol{\Lambda}}$ is a positive diagonal matrix, the entries of its reciprocal square root $\widehat{\boldsymbol{\Lambda}}^{-\frac{1}{2}}$ are the reciprocals of the positive square roots of the respective entries of $\widehat{\boldsymbol{\Lambda}}$. Each diagonal value in the degree matrix $\widehat{\boldsymbol{\Lambda}}$ counts the number of edges connecting the respective node in the graph described by $\widehat{\boldsymbol{A}}$. Such transformation from $\boldsymbol{A}$ to $\widehat{\boldsymbol{A}}$ is in accord with our choice of distance-partitioning [17], where each neighbor set is divided into two subsets for weight assignment, namely the center node ($\boldsymbol{I}$) and the neighbor nodes ($\boldsymbol{A}$). $\mathbf{f}_{in}^{GC}$ is the input feature matrix, and $\mathbf{f}_{in}^{GC} = \boldsymbol{X}_t^{\mathcal{G}}$ at the first layer of input level. $\boldsymbol{W}$ is the layer-wise weight matrix. We refer readers to the appendix section for a more detailed description about graph convolution.

#### *3.1.4 Connecting Graph Convolution with LSTM.*

Here, we describe how the GCN and LSTM layers are connected, as used in both HAR and PBD modules of our hierarchical architecture. For each module, the input to a single unit of the first LSTM layer is the concatenation of the graph convolution output from all the nodes in the graph $\mathcal{G}$ at timestep $t$, denoted by $\mathbf{f}_{out}^{GC}(\boldsymbol{X}_t^{\mathcal{G}}) = [f_{out}^{GC}(v_{t1}), \dots, f_{out}^{GC}(v_{tN})]^{\intercal}$. We want to investigate whether graph representation improves the PBD performance or not, so the GCN and LSTM should not be integrated completely. For the adopted forward-processing LSTM layer, the computation at each LSTM unit is repeated to process the information across graphs from the first timestep to the last. Such conceptually-simple design involving the graph convolution only as a way to learn representations enables us to empirically study its impact on PBD performances. In comparison, others conducted the graph convolution within the gates of each LSTM unit [24] or used extra computational blocks between the GC and LSTM layers (e.g. fully-connected layers used in [18], pooling and attention mechanism applied in [28]).

### **3.2 Hierarchical Connection of HAR and PBD Modules**

Up until this point, the GC-LSTM network used in each module of our proposed architecture has been defined. Here, we describe how to connect HAR and PBD modules. In each module, a fully-connected softmax layer is added to the GC-LSTM network for classification. Let the probability toward each class of the current input frame to be $\boldsymbol{P} = [p_1, \dots, p_K]$ with $K$ denoting the number of classes, with $\boldsymbol{Y}$ being the one-hot prediction. $K$ is 6, including the 5 AoIs and transition activity class for the HAR module, and is 2 for protective and non-protective behavior of the PBD module. In our proposed architecture, to provide activity-informed input from HAR to PBD, a node-wise concatenation is used where the predicted activity label $\boldsymbol{Y}^{HAR}$ is added to the input matrix $\boldsymbol{X}_t^{\mathcal{G}}(v_{ti}) = [x_{ti}, y_{ti}, z_{ti}]$ of each node of the graph input for PBD (see Figure 2). Namely, for the PBD module, activity-informed input feature matrix at a node $v_{ti}$ of a single graph is $\boldsymbol{X}_t^{\mathcal{G},PBD}(v_{ti}) = [\boldsymbol{X}_t^{\mathcal{G}}(v_{ti}), \boldsymbol{Y}^{HAR}]^{\intercal}$. Since the raw graph input fed to the PBD module is joined by the output of the HAR module, we call such a **hierarchical connection** between the two.

### 3.3 Addressing Class Imbalances during Training with CFCC Loss

A problem with datasets simulating real-life situations is class imbalance (e.g. datasets for HAR [21-23]). In the case of the EmoPain dataset, protective behavior is sparsely spread within the AoIs of each data sequence, while it is generally absent during transition activities (see Figure 1). Specifically, on average the AoIs represent only 31.71% of a participant's data sequence, with the rest being transition activities. Furthermore, on average, samples labelled as protective behavior represent only 21.09% of a patient's data sequence, with the rest labelled as non-protective. Typical approaches used to address class imbalance include: i) data re-sampling for each class, where samples are either duplicated from the less-represented class or randomly sampled from the majority class [27]; ii) loss re-weighting, e.g. setting higher weights for the less-represented class and lower weights for the majority class [26]. Unfortunately, these require interferences with data samples that could also harm the training of a model [60].

In our work, we propose to use a loss function that directly alleviates class imbalance during training. Normally, for the supervised learning of our modules, the following categorical cross-entropy loss (CCE) [66] is used

$$\mathcal{L}_{categorical}(\boldsymbol{P}, \boldsymbol{Y}) = -\boldsymbol{Y}\, log(\boldsymbol{P}), \qquad (2)$$

where $\boldsymbol{P} = [p_1, \dots, p_K]$ is the predicted probability distribution of an input frame over the $K$ classes, and $\boldsymbol{Y}$ is the respective one-hot categorical ground truth label with $\boldsymbol{Y}(k) = 1$ only for the ground truth class $k$. During training, the loss computed for each frame is added up to be the total loss for the model to reduce. Such function tends to bias the model to put more attention on decreasing the loss in the majority class and ignores the (mis)classification of the less-represented classes (e.g. the AoI classes in the HAR or the protective behavior class in the PBD task).

To address this problem, we took inspiration from the research on automatic object detection. In object detection domain, a binary-class imbalance exists given the smaller area covered by the object-of-interest and the larger objectless background. Two main approaches proposed in this direction are the focal loss [59] and the class-balanced term [60]. Based on binary cross-entropy loss [66], focal loss applies a **sample-wise** factor function adjusting the loss weight for a sample based on its classification difficulty (judged by the predicted probability towards the ground truth class). The focal loss (FL) together with binary cross-entropy loss (CE) can be written as

$$FL(p, y) = (1 - p_{GT})^{\gamma} \mathcal{L}_{binary}(p, y) = -(1 - p_{GT})^{\gamma} (y log(p) + (1 - y) log\,(1 - p)), \quad (3)$$

where $p$ is the predicted probability towards the positive class of the current data sample, $y$ is the binary ground truth indicator with 1 for the positive class and 0 for the negative class, $p_{GT} = yp + (1 - y)(1 - p)$ is the predicted probability towards the ground truth class. As we can see, the factor $(1 - p_{GT})^{\gamma}$ with tunable hyper-parameter $\gamma \geq 0$ is added to the original binary cross-entropy loss. The intuition is to reduce the loss computed from data samples that are well-classified, while the threshold for judging this needs to be tuned given different datasets and is controlled by $\gamma$. The increase of $\gamma$ will reduce the threshold, then data samples with comparatively lower classification probabilities toward the ground truth class would be treated as the well-classified.

In [60], the authors further revised the vanilla cross-entropy loss by adding a **class-wise** loss weight to each class based on the so-called effective number of samples within it. For class $c$, the effective number of samples is denoted as $E_{n_c} = \frac{1-\beta^{n_c}}{1-\beta}$, with a hyper-parameter $\beta$ controlling how fast the effective samples number $E_{n_c}$ grows when the actual number of samples $n_c$ increases. The class-balanced term is then the reciprocal of $E_{n_c}$ written as

$$\frac{1}{E_{n_c}} = \frac{1-\beta}{1-\beta^{n_c}}. \qquad (4)$$

Unlike the binary imbalance caused by the area of object and its useless background, in the HAR module, class imbalances exist among the 6 categories of activity, while in PBD both protective and non-protective classes share the same importance. Therefore, to adapt the focal loss and class-balanced term to scenarios of HAR and PBD, we replace the CE with CCE and combine the Equation 2-4 as

$$\mathcal{L}_{CFCC}(\boldsymbol{P},\boldsymbol{Y}) = -\frac{1-\beta}{1-\beta^{n_k}}(1-\boldsymbol{YP})^{\gamma}\boldsymbol{Y}\,log(\boldsymbol{P}), \qquad (5)$$

where $n_k$ is the number of frames of the ground truth class $k$ for the current input frame. This revised function, referred to as **C**lass-balanced **F**ocal **C**ategorical **C**ross-entropy (CFCC) loss, will be used in our study. To the best of our knowledge, this is the first time for such a combination to be used for the computation of multi-class categorical cross-entropy loss in HAR and PBD. With CFCC loss, we aim to alleviate class imbalances during training and also to understand its impact in comparison with the other component of our architecture.

## 4 EXPERIMENT SETUP

In this section, we describe more details about the dataset, validation method, metrics, and model implementations.

### 4.1 The EmoPain Dataset and Data Preprocessing

The EmoPain dataset [32] used in this study contains movement data collected from 18 IMUs of 12 healthy and 18 CP participants. The placements of IMUs are illustrated in Figure 3(a). Four wireless surface electromyographic sensors (sEMG) were also used and placed on the high and lower corners on the back of a participant to capture the muscle activity. In this paper, we focus on the movement data and leave the exploration of muscle activity to future work. As part of the EmoPain dataset, the annotation of protective behavior was provided by four domain-expert raters, including 2 physiotherapists and 2 clinical psychologists. Each expert rater independently inspected the on-site video of each CP participant that was collected in synchrony with the wearable sensor data. They marked the timesteps where each period of protective behavior started and ended. The healthy participants were assumed to show no protective behavior despite may having their own idiosyncrasies.

A sequence of functional activities was designed by physiotherapists, comprising one-leg-stand, reach-forward, stand-to-sit, sit-to-stand and bend-down. These activities were selected to reflect the physical and psychological capabilities necessary for carrying out daily functioning. For instance, a person may need to **reach forward** to take an object placed on the far-end of a table, or **bend down** to load the wash machine. They can be also considered as the building blocks for more complex functional activities. For example, reach-forward is usually adopted during kitchen cleaning. Therefore, the HAR experiment conducted in this work shall shed light into relevant HAR studies. Avatar examples of healthy and CP participants conducting the five AoIs are shown in Figure 4. The figure illustrates several strategies used by the CP participant as forms of protective behavior: i) unwilling or unable to raise up the leg during one-leg-stand; ii) avoidance of bending the trunk during reach-forward and bend-down; iii) hesitation, trunk twisting and shoulder side inclination for arm support during stand-to-sit and sit-to-stand;

Each participant went through at least one trial of the activity sequence (~10 mins), while 5 healthy and 11 CP participants executed both the normal and difficult trials. As a result, we have a total of 46 activity sequences from 30 participants. During the **normal** trial, participants were free to perform these activities without any constraint. In the **difficult** trial, they were required to start the activity under the instruction of the experimenter and carry an extra 2Kg weight in each hand during reach-forward and bend-down to simulate picking up daily objects (e.g. shopping bags as typically suggested by physiotherapists). Such difficulty is added to collect the movement of a participant under external pressure or needs. However, no instructions of how a movement should be performed were provided, to ensure participants performed the activity in their own way as what they would do in real life. Between trials or activities, they were allowed to take break as needed and in the way they felt useful to relax or decrease muscle tension. The annotation of activity type was conducted manually by the EmoPain researchers, defining the starting and ending timesteps of each individual activity.

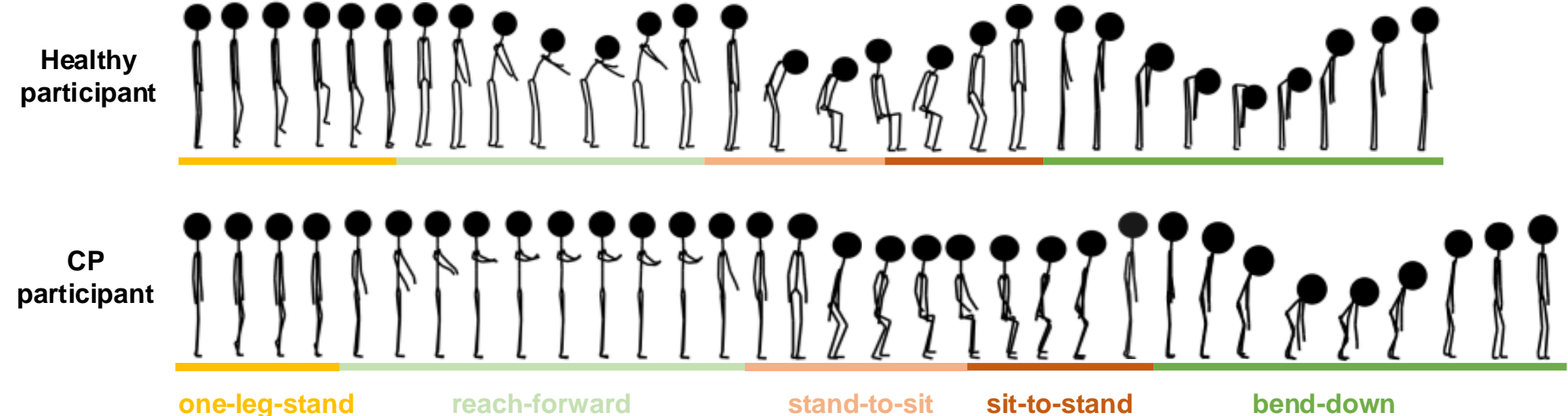

Figure 4: Avatar examples of 3D joint coordinates data from healthy and CP participants. The length of each activity is only the approximation of its real duration.

#### *4.1.1 3D Joint Coordinates as Input.*

Since the basic processing component of our proposed architecture operates on skeleton-like graphs, 3D coordinates of the 22 joints calculated from the18 IMUs are directly used. This differs from previous studies [25, 36, 37, 40, 64] on the same dataset, where low-level features as energies and angles between body segments were used. As described in Section 3.1.1, the 3D joint coordinates were calculated from the BVH data returned by the motion capture suit (Animazoo IGS-190 [81]) using a Matlab MoCap toolbox [70]. The BVH data comprises the skeleton proportion, sensor placements, accelerometer, and gyroscope data sequences recorded at 60Hz.

#### *4.1.2 Continuous Data Segmentation with Sliding-window.*

Using a sliding window of 3s long and 50% overlapping ratio, each activity sequence of a complete trial of a participant is extracted into consecutive frames from the start of the first AoI to the end of the last AoI/transition activity. The window length and overlapping parameters are based on the evaluation studies reported in [25, 40]. At timestep $t$, we have an input graph $\mathcal{G}_t = (V_t, E_t)$, represented by the input data matrix $\boldsymbol{X}_t^{\mathcal{G}}$, constant adjacency matrix $\boldsymbol{A} \in \{0,1\}^{22\times22}$, and its identity matrix $\boldsymbol{I}_{22}$, where $\boldsymbol{X}_t^{\mathcal{G}}(v_{ti}) = [x_{ti}, y_{ti}, z_{ti}], v_{ti} \in V_t$. These matrices only represent the graph structure and 3D joint coordinates data of each joint. The graph structure simulating the human body includes the set of nodes and their connections. The activity class ground truth, i.e. one-leg-stand, reach-forward, sit-to-stand, stand-to-sit, bend-down, and the transition, of a frame is defined by applying majority-voting to the 180 samples within it. The protective behavior ground truth of a frame is also decided by majority-voting across the 4 domain-expert raters in accord with [25, 40, 64]. A frame is labelled as protective behavior if at least 50% of the samples within it had been considered as protective behavior by at least two expert raters separately.

#### *4.1.3 Data Augmentation.*

In order to address the limited size of EmoPain dataset, we apply a combined data augmentation approach that has shown clear improvement in performance in a previous work on the same dataset [25], namely *Jittering* and *Cropping* [77]. For jittering, the normal Gaussian noise is globally applied with standard deviations of 0.05 and 0.1 separately to the original data sequence. For cropping, data samples at random timesteps and joints are set to 0 with selection probabilities of 5% and 10% separately. These augmentations are also beneficial for simulating the real-life situations of signal noise and accidental data loss. Each single augmentation method would create two extra augmented data sets, which is only used for the training set. The original number of frames produced with the sliding-window segmentation from all participants is ∼6200, and is increased to ∼31K after the augmentation.

### 4.2 Validation Method and Metrics

For all the experiments, a leave-one-subject-out cross validation (LOSO) is applied. For each testing participant, both the normal and difficult trials (if conducted) are left out. It should be noted that data augmentation is only applied to the training data, while the original data of the participant left out for testing stays untouched. For HAR, we report the accuracy (Acc) and macro F1 score (Mac.F1) to account for performances of all classes [72]. For PBD, as it is a binary task suffering from class imbalance, we additionally use the protective-class classification output of all folds to plot precision-recall curves (PR curves) and report the area-under-the-curve (PR-AUC) [73].

### 4.3 Model Implementations

A search on number of layers, convolutional kernels, and hidden units for the GC-LSTM network is conducted to identify the suitable hyper-parameter set for HAR and PBD modules separately: i) for HAR, we use one graph convolution layer with 26 convolutional kernels, three LSTM layers with 24 hidden units of each, and one fully-connected softmax layer with 6 nodes for output; ii) for PBD, we use three graph convolution layers with 16 convolutional kernels of each, three LSTM layers with 24 hidden units of each, and one fully-connected softmax layer with 2 nodes for output. A dropout layer with probability of 0.5 is added to each graph convolution layer and LSTM layer to alleviate the overfitting risk for all the models. If not mentioned, the default loss used for all the models is the vanilla categorical cross-entropy loss described in Equation 2. In CFCC loss, the class-balanced term does not vary per sample, instead it is acquired for a class given the number of samples therein, so is computed and fixed before network training. Thereon, we further conduct a hyper-parameter search on $\gamma = \{0,0.5,1,1.5,2,2.5\}$ and $\beta = \{0.9991,0.9995,0.9999\}$ for both tasks separately using the respective HAR or PBD module alone. We find $\gamma = 0.5, \beta = 0.9999$ to be suitable for HAR, and $\gamma = 2, \beta = 0.9999$ for PBD. The Adam algorithm [67] is used as optimizer for all the models, while the learning rate is set to $5e^{-4}$ for the HAR module and $1e^{-3}$ for PBD module, after another search on $lr$ $=\{1e^{-5}, 5e^{-5}, 1e^{-4}, 5e^{-4}, 1e^{-3}, 5e^{-3}\}$. The number of epochs is set to 100 for all the models. The OPTUNA [30] framework is used for the hyper-parameter search. During the hyper-parameter search, a hold-out validation is adopted where 11 healthy and 17 CP participants are randomly selected to be the training set, with the rest left out for validation. The validation data is then removed for the respective LOSO experiments. The aim of the hyper-parameter search is to determine a proper set of hyper-parameters to aid the following experiments.

## 5 RESULTS

The evaluation concerns several components of our proposed hierarchical HAR-PBD architecture, namely the use of graph representation, CFCC loss, and the hierarchical architecture connecting HAR and PBD modules. We conclude by evaluating different training strategies of the hierarchical architecture, and its performances under different sizes of the body graph input.

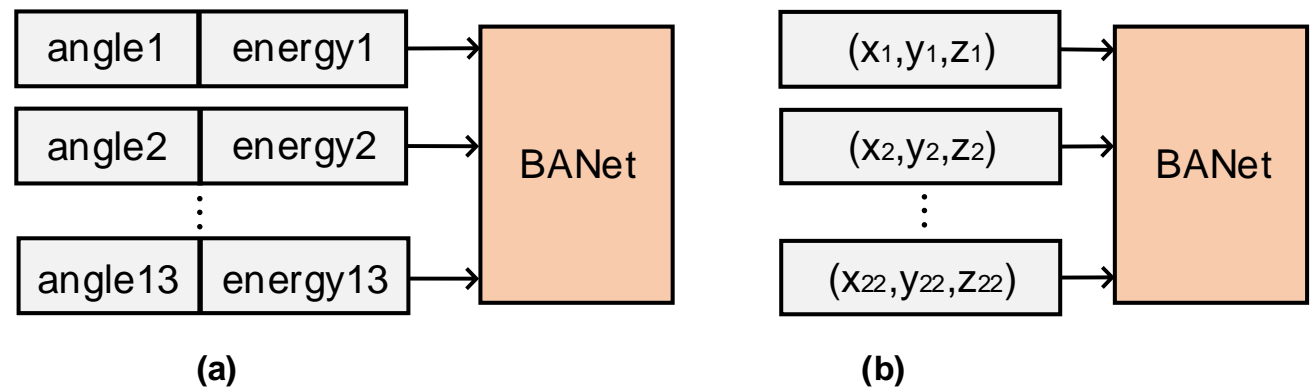

Figure 5: Input structures of a) the original BANet, and b) the adapted BANet for 22 pairs of 3D joint coordinates.

### 5.1 Contribution of Graph Representation to Continuous PBD

The first aim of our evaluation is to understand the contribution of graph representation in comparison with other learning approaches to the PBD performance. Hence, we conduct a set of experiments using the PBD module alone, without the use of the entire hierarchical architecture and CFCC loss. The evaluation is conducted against the stacked-LSTM [40] and BANet [64], which either take i) joint angles and energies; or ii) 22 pairs of 3D joint coordinates as input. For stacked-LSTM, at each timestep we merely concatenate the coordinates of 22 joints to form the input matrix with a dimension of $22 \times 3 = 66$. Accordingly, the input structure of BANet is adapted for 22 pairs of coordinates as illustrated in Figure 5. The search on number of LSTM layers, hidden units, and learning rates is also conducted for the two comparison models respectively under each input condition. Differently from their original studies [40, 64] that relied on pre-segmentation of activity instances, both methods are applied here over the full data sequences in a continuous manner. Results are reported in Table 1 with PR curves plotted in Figure 6. As shown, the PBD GC-LSTM produces the best accuracy of 0.82, macro F1 score of 0.66, and PR-AUC of 0.44. The actual difference between these compared methods is the way the input data is processed with, i.e., traversal concatenation (stacked-LSTM [40]), local processing (BANet [64]), and graph representation (PBD GC-LSTM). As such, the results suggest that the graph representation may indeed contribute to improving the continuous detection of protective behavior. Still, the below-chance-level (<0.5) results of PR-AUC of all methods demonstrate the difficulty of PBD in continuous data sequences. This implies the need to further improve continuous PBD with HAR and CFCC loss.

Table 1: PBD Results of different representation learning methods

| Methods | Acc | Mac.F1 | PR-AUC |
|---|---|---|---|
| Stacked-LSTM (angle+energy) | 0.79 | 0.61 | 0.23 |
| BANet (angle+energy) | 0.78 | 0.56 | 0.24 |
| Stacked-LSTM (coordinate) | 0.80 | 0.64 | 0.32 |
| BANet (coordinate) | 0.79 | 0.63 | 0.27 |
| **PBD GC-LSTM** | **0.82** | **0.66** | **0.44** |

### 5.2 Contribution of CFCC Loss and HAR

Through an ablation study, here we first investigate the contribution of CFCC loss alone in dealing with the imbalanced data for each module of our proposed architecture. We then use our proposed architecture to understand the impact of activity-class information produced by the HAR module on PBD performance. In particular, we aim to understand if recognizing the activity background has more impact on improving PBD in continuous data sequences, in comparison with the issue of class imbalances during training.

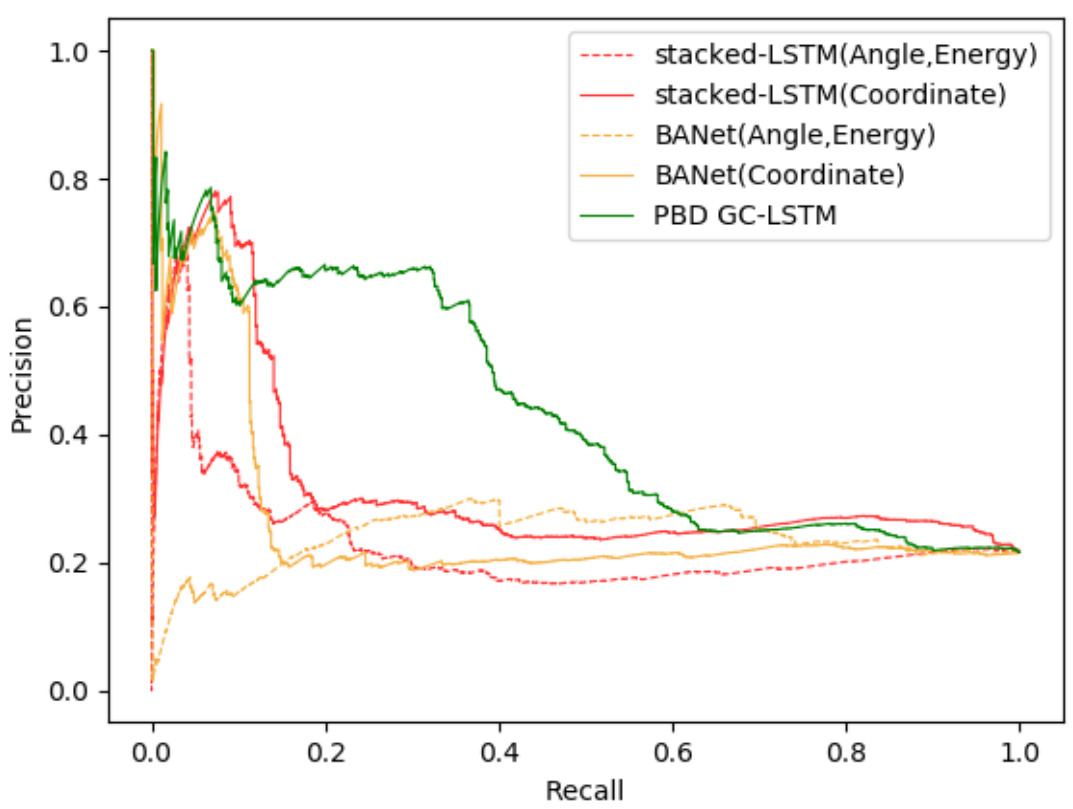

Figure 6: PR curves of different representation learning methods.

#### *5.2.1 Contribution of CFCC Loss to Continuous HAR.*

In our proposed hierarchical HAR-PBD architecture, the HAR GC-LSTM together with CFCC loss was firstly pre-trained on the same set of data using activity labels. Then, the weight achieving the best activity recognition performance was saved and frozen during the training of the entire architecture. For the training and testing of the hierarchical architecture, the HAR output was used as auxiliary information to contextualize the PBD. Therefore, the accuracy of the HAR module is important. Here, we analyze the performance of the HAR GC-LSTM alone with and without CFCC loss. The results are reported in Table 2, with confusion matrices shown in Figure 7. The CFCC loss leads to a higher macro F1 score (0.81 vs. 0.79) in the continuous HAR. Judging from the confusion matrices, CFCC loss reduces the classification bias towards the most represented class (the transition activity), which resulted in a lower accuracy though (0.88 vs. 0.89). These results show the effectiveness of CFCC loss for balancing multi-class categorical loss computation, which was not directly evaluated in the original studies [59, 60]. As is seen in Section 3.3, the computation of CFCC loss is independent of learning models and only requires the prior knowledge of number of samples per class. Therefore, CFCC loss should be useful for HAR tasks on relevant datasets.

Table 2: HAR results of the ablation study

| Methods | Acc | Mac.F1 |
|---|---|---|
| HAR GC-LSTM | 0.89 | 0.79 |
| **HAR GC-LSTM with CFCC loss** | 0.88 | **0.81** |

#### *5.2.2 Contribution of CFCC Loss to Continuous PBD.*

Here we investigate the contribution of CFCC loss to continuous PBD using the PBD GC-LSTM. The input to PBD GC-LSTM is the 3D joint coordinates data without activity-class information. As we can see from the results in Table 3, the use of CFCC loss leads to ~5% improvement in macro F1 score (macro F1 score of 0.71 vs. 0.66). Confusion matrices shown in Figure 8(a)(b) suggest that the adapted CFCC loss does indeed help penalize the bias towards the more frequent class (non-protective class in this case) while improving the recognition of the less-represented one (protective class). However, the PR-AUC of 0.48 is still below chance level, suggesting that addressing class imbalance alone is not sufficient.

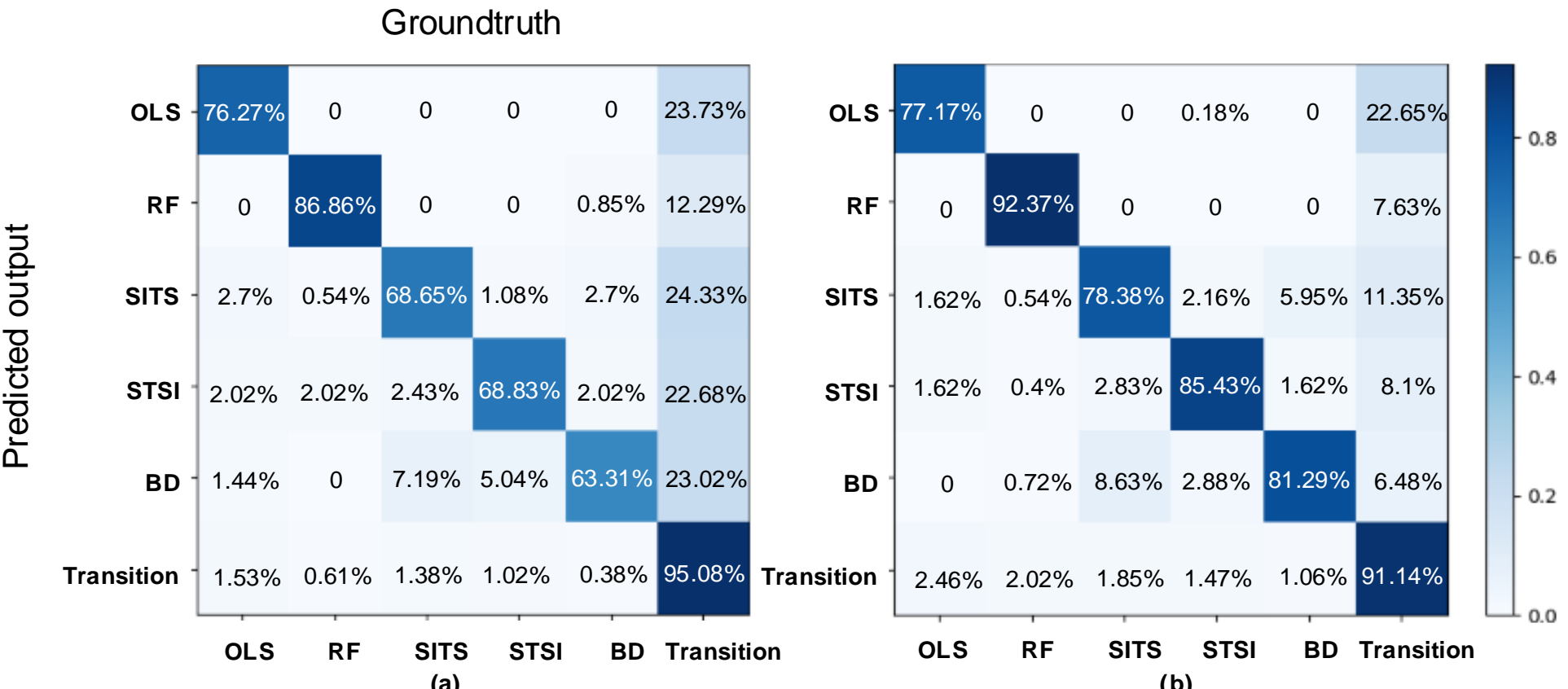

Figure 7: Confusion matrices of a) HAR GC-LSTM and b) HAR GC-LSTM with CFCC loss. OLS=one-leg-stand, RF=reach-forward, SITS=sit-to-stand, STSI=stand-to-sit, and BD=bend-down.

#### *5.2.3 Contribution of Hierarchical HAR-PBD Architecture to Continuous PBD.*

For the training and testing of our proposed hierarchical HAR-PBD architecture, the HAR GC-LSTM within it is frozen and loaded with the weights from its pre-training with CFCC loss. This is to keep the HAR performance constant and aid the understanding of the impact of continuously inferred activity information on continuous PBD. The results are reported in Table 3, with confusion matrix shown in Figure 8(c). It is interesting to see that our proposed hierarchical HAR-PBD architecture using vanilla categorical cross-entropy loss achieved an improvement of ~2% with respect to the PBD GC-LSTM alone using CFCC loss (macro F1 score of 0.73 vs. 0.71). The PR-AUC of 0.52 is also above chance level. Such result shows that the contextual information of activity type contributes to continuous PBD with our proposed hierarchical HAR-PBD architecture being a practical way to leverage such information. Furthermore, by adding CFCC loss to the PBD module of the hierarchical HAR-PBD architecture, higher macro F1 score of 0.81 and PR-AUC of 0.60 are achieved (confusion matrix shown in Figure 8(d)). The PR curves for the PBD ablation study are plotted in Figure 9. These results highlight that the contextual information of activity types played a higher role in improving PBD in continuous data, while adding a mechanism (CFCC loss in our case) to address the class imbalance problem led to a further-clear improvement. Such suggest that both the HAR and CFCC loss are necessary for continuous PBD despite one being more effective than the other.

### 5.3 Comparing Training Strategies of the Hierarchical Architecture

In the previous subsections, the HAR module used in hierarchical HAR-PBD architecture was pre-trained with the same training data using activity labels and frozen to adopt the model of best activity recognition performance. The aim was to understand the contribution of HAR to PBD across the different configurations. Here, we further explore the relationship between HAR and PBD modules by exploring joint-training strategies of the hierarchical architecture. In joint-training of the architecture, the HAR module would not be frozen but the activity labels are still used to update it when the PBD module is trained. Specifically, the protective behavior labels of the same data input together with the output of HAR module are used to train the PBD module. Thereon, we compare the following four joint-training strategies together with the use of CFCC loss:

1. ***Joint HAR(CFCC)-PBD*** and ***Joint HAR-PBD(CFCC)***, where HAR and PBD modules are initialized and trained together using activity and protective behavior labels respectively, with CFCC loss only added to either the HAR or PBD module;

2. ***Joint HAR-PBD with CFCC***, where CFCC loss is added to both modules in such joint training;
3. ***Pre-trained Joint HAR(CFCC)-PBD*** and ***Pre-trained Joint HAR-PBD(CFCC)***, similar to (1) where the only difference is that the HAR module is first trained alone with activity labels using CFCC loss to achieve the best activity recognition performance and then its training continues with the training of the PBD module;
4. ***Pre-trained Joint HAR-PBD with CFCC***, where CFCC loss is added to both modules in the joint training of (3).

Table 3: PBD results of the ablation study

| Methods | Acc | Mac.F1 | PR-AUC |
|---|---|---|---|
| PBD GC-LSTM | 0.82 | 0.66 | 0.44 |
| PBD GC-LSTM with CFCC loss | 0.83 | 0.71 | 0.48 |
| Hierarchical HAR-PBD architecture | 0.84 | 0.73 | 0.52 |
| **Hierarchical HAR-PBD architecture with CFCC loss** | **0.88** | **0.81** | **0.60** |

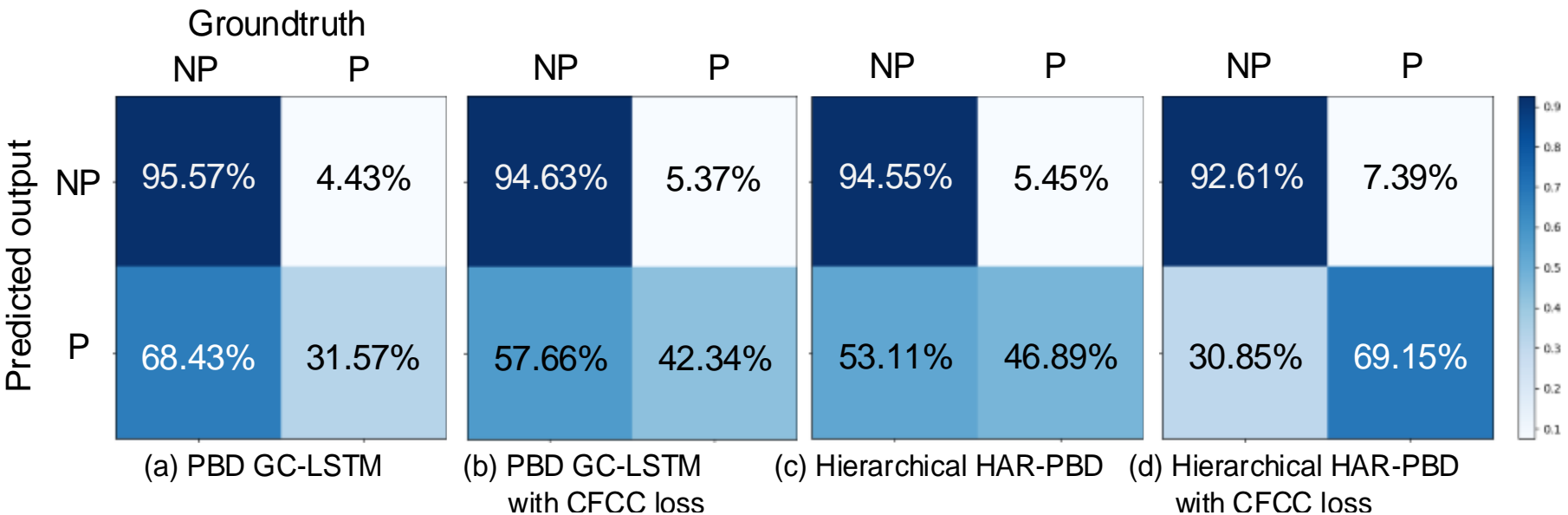

Figure 8: Confusion matrices for PBD methods in the ablation study. NP= non-protective, P=protective.

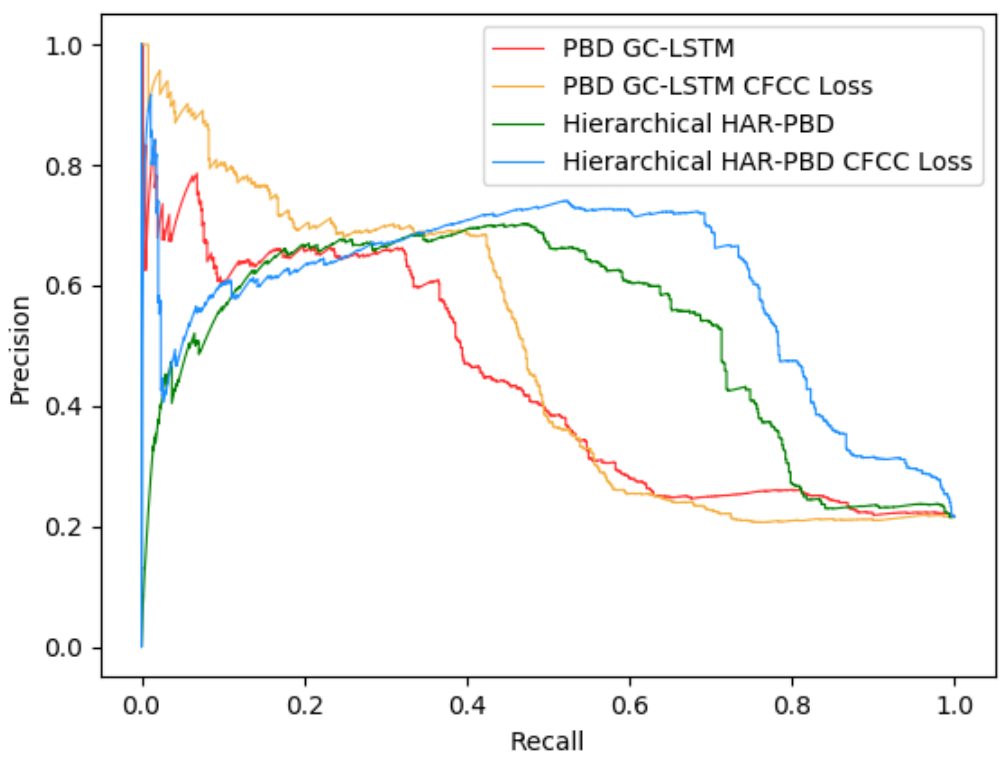

Figure 9: PR curves of different PBD methods in the ablation study.

For all these joint training strategies, the loss weights are set to {1.0, 1.0} for both HAR and PBD modules. If CFCC loss is not mentioned, the loss used for the respective module is the vanilla categorical cross-entropy loss described in Equation 2. We also compare them with our default method used in previous sub-sections, here referred to as ***Pre-trained HAR(Frozen)-PBD(CFCC)***, where the HAR module is first trained alone with activity labels and CFCC loss to achieve the best activity recognition performance per LOSO fold, then it is **frozen** with weights loaded and used in the hierarchical architecture for training and testing of the PBD module. Results are reported in Table 4, with the PR curves for PBD results plotted in Figure 10(a).

Without pre-training the HAR module, the best HAR (macro F1 score of 0.56) and PBD (macro F1 score of 0.74 and PR-AUC of 0.55) performances are achieved by the *joint HAR-PBD(CFCC)*. However, by adding CFCC loss to the HAR module alone (*joint HAR(CFCC)-PBD*), the performances are reduced notably for the HAR and slightly for PBD. One explanation could be that, the error passed back from the PBD module harmed the HAR performance, especially when such error of PBD was not well handled e.g. without using CFCC loss. On the other hand, by adding CFCC loss to both modules (*joint HAR-PBD with CFCC*), the HAR performance achieved (macro F1 score of 0.54) is comparable to *joint HAR-PBD(CFCC)* but the PBD performance is much lower (macro F1 score of 0.71 and PR-AUC of 0.45). Given the current hierarchical architecture, such results suggest that alleviating class imbalance in PBD has a stronger impact on the overall performance in joint training, while addressing it in HAR penalizes the PBD performance.

Rather than to start joint training from scratch, we further look into the uses of pre-training of the HAR module to reach an initial best activity recognition performance (macro F1 score of 0.81) before joint-training. A similar outcome as above is observed where the best performance is achieved by adding CFCC loss to the PBD alone. Once again this proved the higher impact of alleviating the class imbalance of PBD, as the error passed back from the PBD module could harm the training of HAR module. In general, the results show that a pre-training of the HAR module improved the final performances of both HAR and PBD modules in comparison to the ones without it.

Table 4: HAR and PBD Results for different training strategies of the Hierarchical HAR-PBD architecture

| | HAR | | PBD | | |
|---|---|---|---|---|---|
| Training strategies | Acc | Mac.F1 | Acc | Mac.F1 | PR-AUC |
| Joint HAR(CFCC)-PBD | 0.62 | 0.42 | 0.85 | 0.70 | 0.54 |
| **Joint HAR-PBD(CFCC)** | 0.76 | 0.56 | 0.84 | 0.74 | 0.55 |
| Joint HAR-PBD with CFCC | 0.66 | 0.54 | 0.81 | 0.71 | 0.45 |
| Pre-trained Joint HAR(CFCC)-PBD | 0.68 | 0.55 | 0.85 | 0.74 | 0.58 |
| **Pre-trained Joint HAR-PBD(CFCC)** | 0.84 | 0.73 | 0.87 | 0.79 | 0.58 |
| Pre-trained Joint HAR-PBD with CFCC | 0.72 | 0.64 | 0.85 | 0.76 | 0.55 |
| **Pre-trained HAR(Frozen)-PBD(CFCC)** | **0.88** | **0.81** | **0.88** | **0.81** | **0.60** |

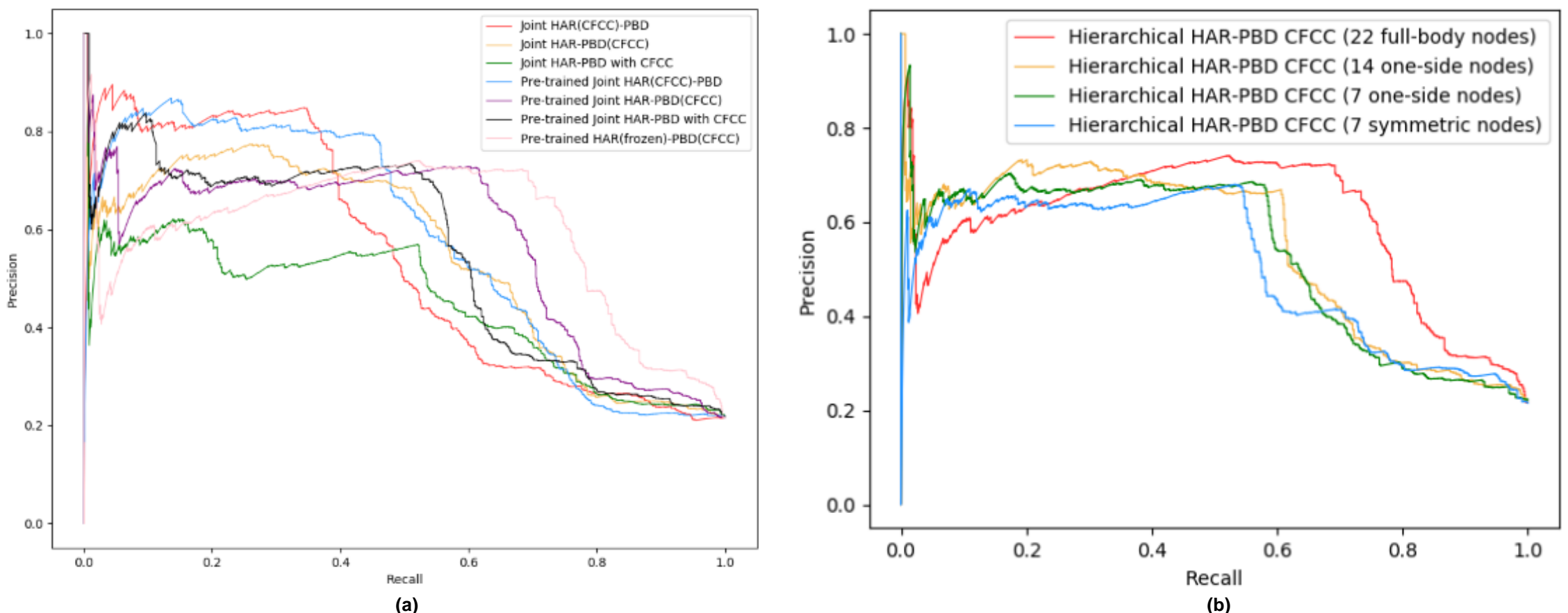

Figure 10: PR curves of the a) hierarchical architecture under different training strategies; b) hierarchical architecture with CFCC loss using input of different sensor sets.

The performances achieved by the various joint-training strategies of the hierarchical architecture are still lower than the one of freezing the HAR module as used in previous sub-sections, for both HAR (macro F1 score of 0.81) and PBD (macro F1 score of 0.81 and PR-AUC of 0.60). It should be noted that this method is a **two-stage** process in training and an **end-to-end** process in inference. Although these results highlight the importance of HAR performance to PBD, they also suggest that the error propagated back from the PBD module in joint-training was not informative to improve the HAR performance. This highlights the need to investigate the interaction scheme between HAR and PBD modules, beyond error back-propagation. We leave such an exploration to future work.

### 5.4 Simulating Fewer IMUs

Until this point, we have assumed all 18 IMUs to be available to enable the input of a full-body graph. In this experiment, we quantify the fluctuation in performance when fewer IMUs are available. We simulate the limited availability of IMUs by removing nodes (containing the data of respective joints) from the full-body graph. According to the study on human observation of protective behavior [36], protective movement strategies are often visible on both sides of the body even if via different patterns. For example, a twisting of the trunk to reach for a chair may lead to a narrower angle between the arm and the trunk on one side but a compensatory-larger angle between another arm and the trunk. Therefore, a ***one-side sensor set*** of 14 nodes is created, where nodes number of 2-4 and 10-14 on the left limbs of the full-body graph are removed. Second, to simulate an even more compact sensor set, we further remove nodes number of 6, 8, 15, 17, 18, 20, and 21 from the one-side sensor set, resulting in a ***smallest one-side sensor set*** of 7 nodes. Additionally, from the full-body graph, we symmetrically remove nodes number of 3, 4, 6, 7, 8, 10-13, 15-18, 20, and 21 from both body sides to create a ***smallest symmetric sensor set*** with **7** nodes as well. The graph structures of these sensor sets still simulate human body connections, as shown in Figure 11. The hierarchical HAR-PBD architecture with CFCC loss is used here on the graph input extracted from each sensor set. For a fair comparison, we conducted another search to determine the suitable hyper-parameters under each condition, with the process detailed in Section 4.3. The HAR and PBD results of each sensor set are shown in Figure 12, with PR curves for the PBD results plotted in Figure 10(b).

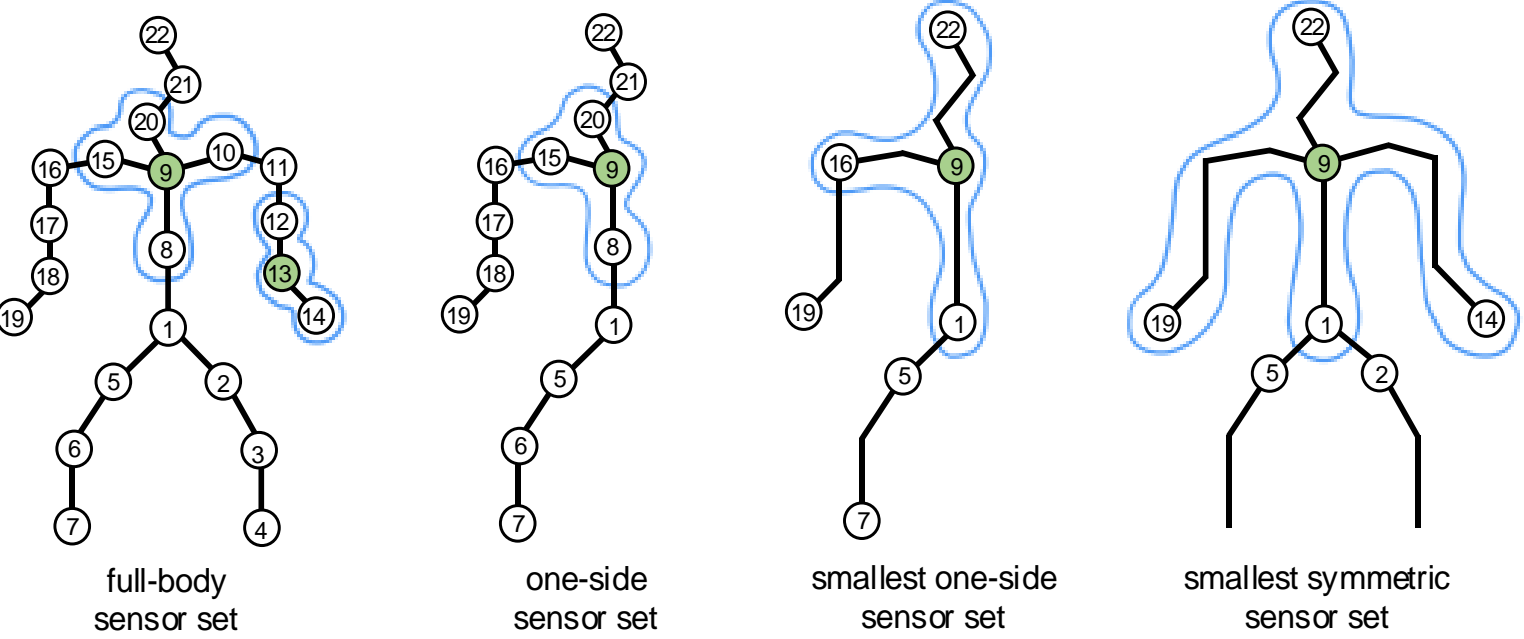

Figure 11: Graph structures of different sensor sets. The blue contour marks the neighbor set of each centered node in green.

Although the best PBD performance is obtained by using the default graph input of 22 nodes (macro F1 score of 0.81 and PR-AUC of 0.60), competitive results are achieved using the one-side graphs with number of nodes reduced to 14 (macro F1 score of 0.77 and PR-AUC of 0.55) and even 7 (macro F1 score of 0.76 and PR-AUC of 0.53). These results are better than the ones achieved using the hierarchical architecture alone without CFCC loss on the full-body graph (macro F1 score of 0.73 and PR-AUC of 0.52). On the other hand, given the same number of 7 nodes, the worst performance is achieved by the smallest symmetric sensor set that follows a general practice of retaining nodes on both sides of the body (macro F1 score of 0.75 and PR-AUC of 0.51). This shows advantage of using an observation-driven strategy in guiding the sensor-set reduction, in the context of PBD. Generally, it is empirically verified that the proposed hierarchical HAR-PBD architecture with CFCC loss leads to improvement even with small sensor sets. In order to further improve the PBD performance, efforts could be made on i) designing better graph structure, since in this work we merely employed the human-body connections; ii) further exploring the configurational pattern of body movement in the context of CP rehabilitation, given the performance achieved by one-side sensor sets.

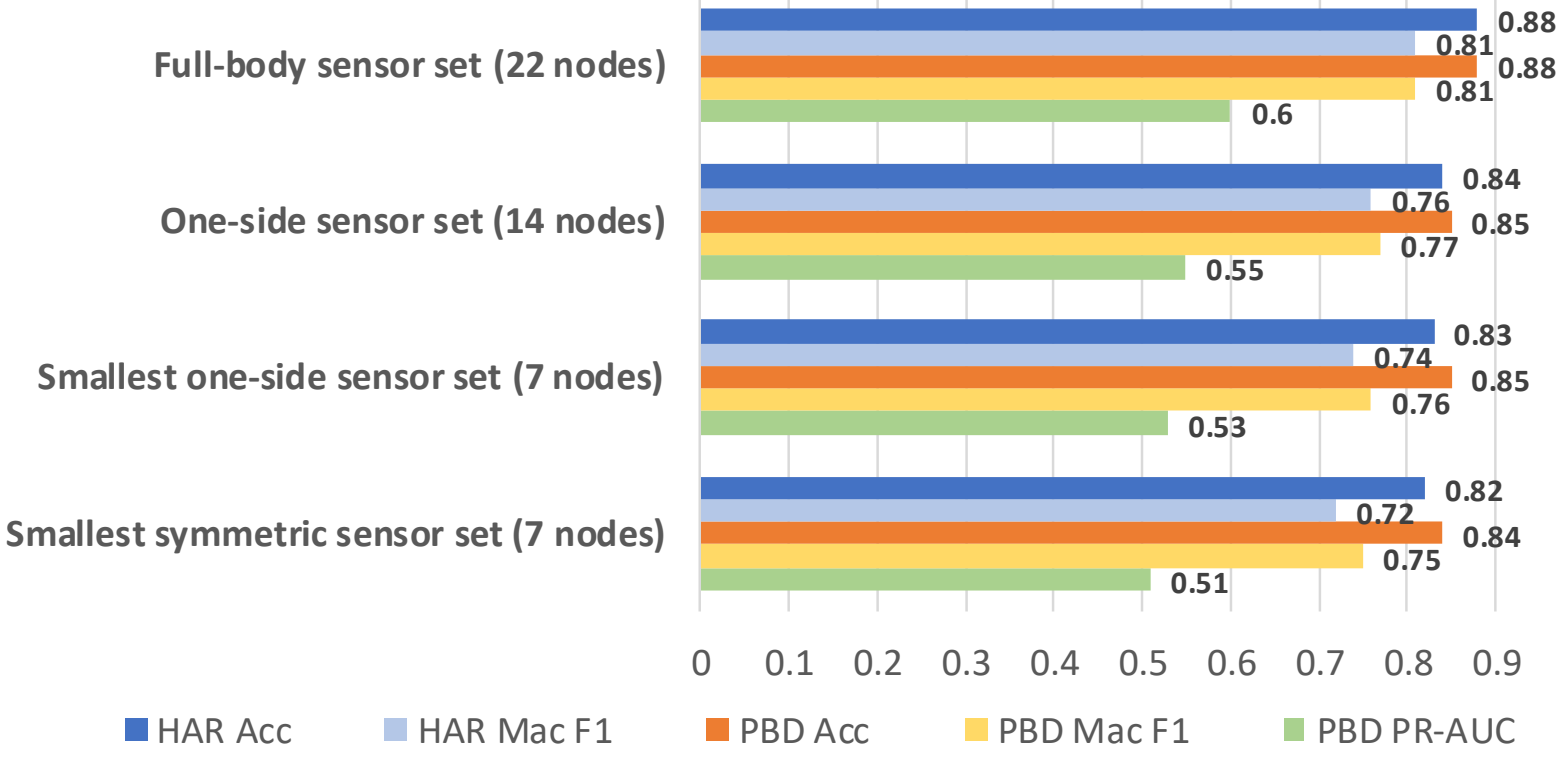

Figure 12: HAR and PBD results of the hierarchical HAR-PBD architecture with CFCC loss using input of different sensor sets.

## 6 DISCUSSION

We discuss here our open challenge, current limitations, and possible use cases of the proposed method.

### 6.1 The Challenge and Current Limitations

The major challenge for ubiquitous-computing research in chronic-pain management is the lack of very large datasets. This is a problem we face as moving into real-world applications. The EmoPain dataset [32] comprises 18 CP and 12 healthy participants, the data collection and annotation of which took nearly a year to finish according to the authors. It is widely acknowledged that collecting data from patients is challenging given increasingly strict data protection regulations and privacy issues. In order to fully leverage the existing data for our model development, we followed the experience of previous studies [25, 40, 64] to use data augmentations [77]. As wearable technology become easier to use in everyday scenario, we plan to conduct a long-term data collection in the future. Other limitations of this work with possible solutions for the future are summarized as follows.

#### *6.1.1 The Dependence on Manual Annotation.*

Our proposed architecture is a pure supervised-learning method that relies on manual annotations, particularly domain-expert ratings of behavior. With expert annotation of protective behavior, labelling frames by majority voting can be problematic, possibly biasing the model in favor of part of the experts. In future work, we would treat this as a noisy-label problem, and model each expert's annotation separately while gaining better consensus via a multi-expert architecture [69]. On the other hand, the annotation of activities is challenging given the variety in performances of CP participants. Particularly, discriminating the margin between activity-of-interest and transition is often difficult and may lead to the misclassification of transition toward AoIs, as reported in Appendix A.2. Our current practice of using majority voting is also unable to deal with the situation when multiple activities exist on the same frame. For this problem, we plan to follow the recent progress in ambiguous activity annotation [31] to drive our future research.

#### *6.1.2 Limited Interaction between HAR and PBD Modules.*

Our experiments show that the HAR module improves PBD, while the error propagated back from the PBD module is not that informative to refine the HAR module. Therefore, aside from simple error back-propagation, one could establish a better interaction scheme between the two. Thereon, a factor could be considered is the granularity of protective behavior type. In this paper, the five typical classes of protective behavior (guarding, hesitation, the use of support, abrupt motion, and rubbing) [33] were pulled together and modelled as one unique class, given the limited number of instances per each type. If modelled separately, these may add more insight to the type of activity being performed. For instance, support is used more during bend-down or stand-to-sit. Hence, new data collection for similar applications should consider how to increase the number of instances per behavior type.

#### *6.1.3 The Use of Large IMUs Network.*

For most experiments in this work, a set of 18 IMUs was assumed to be available to provide data of the full-body graph (22 joints). So many IMUs are not usually directly taped to the body, and we do not expect such to be the case when the system is deployed. In fact, ubiquitous motion capture suits that facilitate sensor wearability, e.g. the Animazoo IGS-190 [81] (used for the EmoPain dataset) and Xsens MVN [82], have been around for a long time. Two examples of the user wearing the MoCap suit are shown in Figure 13. Both systems are integrated, wireless, and consider users' comfort. However, such motion capture systems are still expensive even though the IMU sensors are becoming cheaper, more accurate, and wearable (e.g. invisible, washable, or transferable between clothes [76]). Currently, it is out of the scope of this paper to develop the suit or integrate sensors into patients' clothes. Still, this remains an open area for hardware developers and fashion designers to propose better solutions. Progress in ubiquitous computing (as the one in our work) may lead to further advances in hardware development, a very active area e.g. the integration of multiple sensors in sport garments. Studies with clinicians and patients show that such advancement is very desirable to manage the conditions [83, 84]. Hopefully, our research may further augment such wearable devices with PBD capabilities and extend applications to rehabilitation and clinical contexts. The original aim of this work is, with a large set of sensors, to understand what is feasible and then explore how to improve it. Several studies have recently aimed to combine sparse IMUs or just accelerometers (less than 6 sensors) and visual clues to reconstruct full-body motions [85, 86]. Given the highest performance is achieved by using full-body graphs, we can follow this work to simulate full-body movement data using a smaller sensor set.

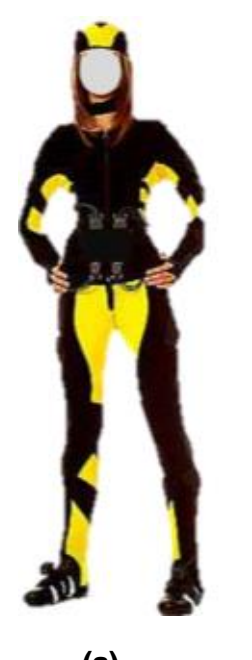
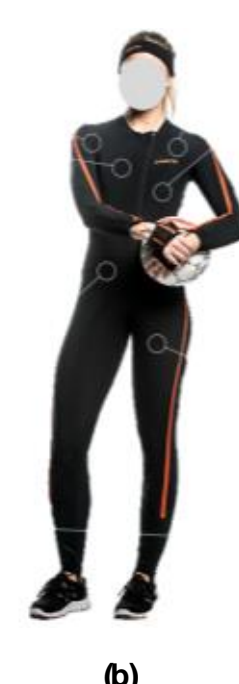

Figure 13: Example of users wearing the a) Animazoo IGS-190 and b) Xsens MVN motion capture suits.

### 6.2 Future Use Cases

While the goal of our study was not to build a ubiquitous support system for pain management, our architecture is a key component of such a system as performance of continuous PBD is critical for effective support. It should be noted that contextualization provided by the HAR module not only leads to improved PBD performance, but informs assessment of people with CP and customizes timely support for self-management. We discuss here the main use cases and further developments that can exploit our proposed hierarchical architecture to deliver new types of support and interventions in CP management and beyond.

#### *6.2.1 In-the-wild-informed Clinical Rehabilitation.*

Clinicians need to know about patients' difficulties in everyday activity [68], beyond the safe environment of the clinic, and without reliance on self-reported behaviors (e.g., diaries) that are commonly used but of low reliability [29] since awareness of habitual protective behavior and their triggers is low [62]. A ubiquitous system, capable of recognizing activity context and continuously detecting protective behavior, can provide clinicians with better understanding of the patient's activity difficulties, and of progress, which often varies across AoIs. Connected to GPS and time, the system could further contextualize the activity, with factors that add stress, e.g. social pressures.

#### *6.2.2 Patient-oriented Ubiquitous Self-management.*

Difficulty transferring movement strategies learned in the clinic to everyday life is common, because of the complexity of the real world (environment, social demands, variety of activities, and responsibilities, etc.), and interference by emotional states [37]. In [29, 79], a ubiquitous system transforms real-time movements (of specific body parts) into sound (sonification) to increase awareness in people with chronic pain of their physical capabilities. This further facilitates the autonomous use of movement strategies of the user beyond the clinic. If integrated in such ubiquitous system, our HAR-PBD architecture could help identify when advice is needed, e.g. when the frequency of protective behavior during specific activities rises above a certain level; it can instantly provide reminders of breathing and breaks as well. Taking breaks and relaxation are critical pacing strategies to avoid tension that could lead to setbacks and prolonged days in bed. During exercise, the system can also provide dedicated suggestions or exercise plans based on the frequency of protective behavior detected.

#### *6.2.3 Beyond Chronic-pain Management to Next-stage Human Activity Analysis.*

Beyond supporting the management of chronic pain, our proposed hierarchical architecture could be applied in a variety of contexts where ubiquitous HAR technology is being leveraged. For example, ubiquitous HAR technology is opening new platform to aid workers in factory assembly lines [15], to support them in their workspace activities, e.g. to identify and help correct mistakes, to aid training, and establish human-robot collaboration. Thereon, another interesting application is to promote workers' wellbeing, such as in reducing mental or physical stress. Our architecture can be integrated into the system to leverage HAR for detecting cues of fatigue or pain. Such a system could help identify the need for a break and adjust working timetables. These are essential to minimize development of musculoskeletal conditions, a common problem in manufacturing industries. In similar contexts, the number of sensors could be reduced to fit the specific activities and relevant movements.

Another active area of application is in healthcare. For instance, in [8], limb movement was assessed to screen perinatal stroke in infants, while arm movement was continuously analyzed to track everyday rehabilitation of stroke patients [75]. For these, integration of our hierarchical architecture in the system could help establish the link between the type of activity/movement and the behavior category (e.g. good or poor rehabilitation engagement for [75], and even pain or anxiety). Such activity-aware functions could allow more in-depth understanding of the patient and generate opportunities for personalized support.

## 7 CONCLUSION

Ubiquitous technologies open new opportunities to support people with chronic pain during their everyday self-directed management. In this paper, we targeted PBD in continuous movement data as the critical first step. We proposed a hierarchical HAR-PBD architecture to recognize the varying context of activity to aid the simultaneous detection of protective behavior. An adapted CFCC loss was also used to alleviate class imbalances of continuous data during training. Our evaluation with data from real patients suggested that the activity type information is effective to aid PBD in continuous data, leading to a notable improvement over the baseline (macro F1 score of 0.73 and PR-AUC of 0.52 vs. macro F1 score of 0.66 and PR-AUC of 0.44), and is more impactful than just solving class imbalances (macro F1 score of 0.71 and PR-AUC of 0.48). The best result was achieved by combining the hierarchical architecture with CFCC loss, with macro F1 score of 0.81 and PR-AUC of 0.60. Additionally, in Section 5.1, we verified that graph representation improves the PBD performance. In Section 5.3, we showed that it is feasible to jointly train the hierarchical HAR-PBD architecture. However, work is needed to gain mutual improvement between HAR and PBD modules. In Section 5.4, we showed the applicability and efficacy of our method using fewer nodes/joints (macro F1 scores of 0.77 and 0.76 with 14- and 7-node data input respectively). In subsequent research, we hope to build on the findings of this paper to establish a ubiquitous CP management system, considering real-world challenges and proper interactions between the user and system.

## A APPENDICES

Here we first describe in detail the graph convolution function used in this paper. Then, we analyze the error of each component of the proposed hierarchical HAR-PBD architecture with CFCC loss via visualization.

### A.1 The Graph Convolution Function

Following the derivation of GCN presented in [17], the GC used in this work can be written in detail as

$$f_{out}^{GC}(v_{ti}) = \sum_{v_{tj}\in\mathcal{N}(v_{ti})} \frac{1}{Z_{ti}(v_{tj})} f_{in}^{GC}(\boldsymbol{p}^{GC}(v_{ti}, v_{tj})) \cdot \boldsymbol{w}^{GC}(l_{ti}(v_{tj})), \qquad (6)$$

where graph-adapted sampling function is $\mathbf{p}^{GC}(v_{ti}, v_{tj}) = v_{tj}$ with $d(v_{ti}, v_{tj}) \leq 1$, the graph-adapted weight function is $\boldsymbol{w}^{GC}(v_{ti}, v_{tj}) = \boldsymbol{w}'(l_{ti}(v_{tj}))$ with $l_{ti}(v_{tj}) = d(v_{ti}, v_{tj})$, and $\boldsymbol{w}'$ to be the trainable weight matrix, $f_{in}^{GC}$ is the input feature of the sampled node set at current layer while $f_{out}^{GC}$ is the output feature of the respective centered node $v_{ti}$, and $Z_{ti}(v_{tj}) = \mathbf{card}(\{v_{tk} | l_{ti}(v_{tk}) = l_{ti}(v_{tj})\})$ is a normalization term representing the cardinality of the partitioned subsets in the neighbor set. The 1-neighbor set $\mathcal{N}(v_{ti}) = \{v_{tj} | d(v_{ti}, v_{tj}) \leq 1\}$ is applied to be the receptive field of each node $v_{ti}$, as depicted by the blue contour in Figure 3(c). Within the weight function, the partition function $l_{ti}: \mathcal{N}(v_{ti}) \rightarrow \{0, \ldots, K-1\}$ can be used under different strategies, while in our work the distance-partitioning strategy [17] is adopted that divides the 1-neighbor set $\mathcal{N}(v_{ti})$ into two subsets, namely the centered node $v_{ti}$ and the remaining neighbor nodes $v_{tj} | d(v_{ti}, v_{tj}) \leq 1$. As a result, we have $K = 2$ subsets thus $l_{ti}(v_{tj}) = d(v_{ti}, v_{tj})$. By using the distance-partitioning strategy, $Z_{ti}(v_{tj})$ equals to the number of all the neighboring nodes $v_{tj}$ within the same neighbor set because they are within the same subset as well.

### A.2 Error Analysis with Visualization

To understand the temporal behavior of the two modules in the hierarchical HAR-PBD architecture, a visualized example of the model performances on the data sequence of one CP participant is shown in Figure 14. The upper two diagrams are the ground truth and recognition result of the HAR module respectively. As shown, on this long data sequence, our HAR GC-LSTM using CFCC loss achieves good performances without any pre-localization and -segmentation of the AOIs. The lower five diagrams are the ground truth and results of the PBD module achieved by the four different methods respectively.

In the HAR result (upper part of figure 14), the errors are found to be: i) misclassification of one-leg-stand as transition activity (red rectangles on the left); ii) misclassification of transition activities as reach-forward and bend-down (red rectangles in the middle); iii) misclassification of bend-down as stand-to-sit (red rectangle in the right). We notice that most misclassified activities were possibly due to their similarity in execution given the use of protective behavior by this CP participant. For instance, the analysis of the on-site recorded video shows that the participant was unable/unwilling to raise the leg up during one-leg-stand, which is similar to the transition activity of standing still. During bend-down, the participant was not to bend the trunk but the leg and reached both arms to the ground, which is similar to the activity of stand-to-sit.

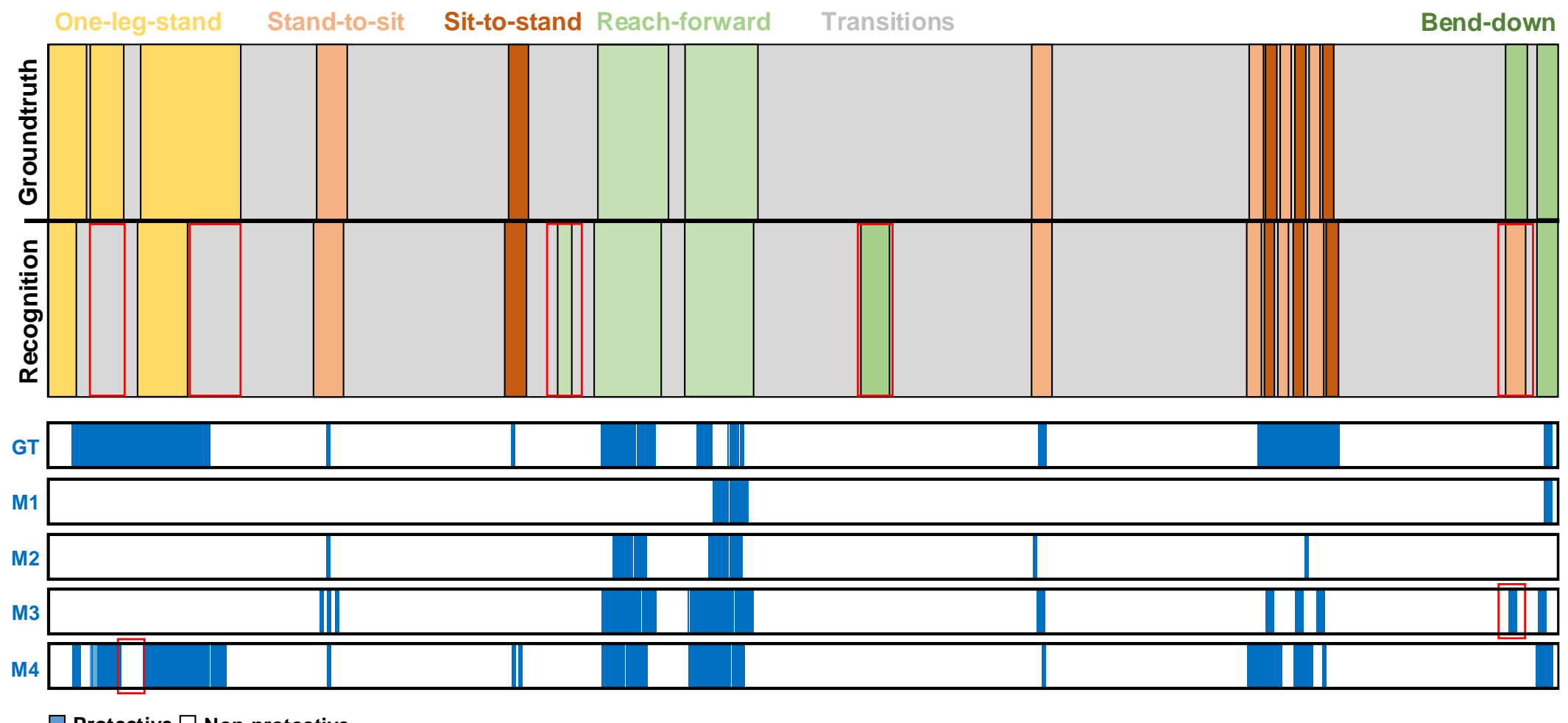

Figure 14: An example of the ground truth and results of HAR and PBD modules for the data of a CP participant. The upper diagram is showing the ground truth of activity class and the recognition result by HAR GC-LSTM with CFCC loss. At the lower diagram, the first row is presenting the ground truth for PBD. 'M1' to 'M4' are respectively the detection result of i) PBD GC-LSTM; ii) PBD GC-LSTM with CFCC loss; iii) hierarchical HAR-PBD architecture, and iv) hierarchical HAR-PBD architecture with CFCC loss.

We now compare the four PBD approaches (see M1-M4 in the lower part of Figure 14). Without the activity-class information and CFCC loss, the baseline PBD GC-LSTM (M1) misclassified most frames as the majority class of non-protective behavior, which takes up around 78.91% in the training data. More protective behavior frames are correctly detected by using CFCC loss (M2), possibly owing to its ability to drive the model to focus more on the less-represented class i.e. the protective behavior class in our case For this CP participant, M3 is shown to be more effective than M2 in terms of PBD during stand-to-sit, sit-to-stand, and bend-down. This could be mainly owed to the activity-type information on these frames provided by the HAR module. The hierarchical HAR-PBD architecture with CFCC loss (M4) leads to the best result, especially for PBD during one-leg-stand. In the PBD result of the hierarchical architecture without CFCC loss (M3), the misclassified area marked by a red rectangle on the right side of the figure seems to be affected by the misclassification of bend-down as stand-to-sit in the HAR module. Such error is corrected by using CFCC loss (M4), possibly because it forces the model to adaptively down-weight the frames of majority class i.e. the non-protective behavior class in our case However, for the same approach (M4), the error marked by a red rectangle on the left side is likely to have been affected by the misclassification of one-leg-stand as transition activity by the HAR module.

These results suggest that i) misclassifications by the HAR module have a negative impact on PBD performance; ii) and this problem could be minimized by addressing the class imbalance with CFCC loss in the PBD module. These support our concept of approaching continuous PBD by addressing the two technical issues together, namely the contextual information of activity types and the imbalanced presence of protective behavior during training.